\pdfoutput=1

\documentclass[11pt]{article}

\usepackage{EMNLP2023}
\usepackage{times}
\usepackage{latexsym}

\usepackage[T1]{fontenc}

\usepackage[utf8]{inputenc}
\DeclareUnicodeCharacter{266B}{}
\usepackage{microtype}

\usepackage{inconsolata}
\usepackage{amsmath}
\usepackage{algorithmicx,algorithm}
\usepackage{graphicx}
\usepackage{subfigure}
\usepackage{caption}
\usepackage{setspace}
\usepackage{mathtools,array}
\usepackage{makecell}
\usepackage{url}
\usepackage{float}
\usepackage{paralist}
\usepackage{threeparttable}
\usepackage[switch]{lineno}
\usepackage{tabularx}
\usepackage{multirow}
\usepackage{booktabs}
\usepackage{graphicx}
\usepackage{algorithm}
\usepackage{algpseudocode}
\usepackage[most]{tcolorbox}

\title{Two-way Evidence self-Alignment based Dual-Gated Reasoning Enhancement}

\author{
Kexin Zhang\textsuperscript{1}\thanks{\ \ Equal contribution.}\quad
Junlan Chen\textsuperscript{1}\footnotemark[1] \quad
Daifeng Li\textsuperscript{1}\footnotemark[1]\thanks{\ \ Corresponding author.} \quad
\textbf{Yuxuan Zhang}\textsuperscript{1} \\
\textbf{Yangyang Feng}\textsuperscript{1} \quad
\textbf{Bowen Deng}\textsuperscript{1} \quad
\textbf{Weixu Chen}\textsuperscript{1} \\ \\
\textsuperscript{1}Sun Yat-sen University, Guangzhou, China \\
\texttt{zhangkx57@mail2.sysu.edu.cn}
}

\begin{document}
\maketitle
\begin{abstract}
Large language models (LLMs) encounter difficulties in knowledge-intensive multi-step reasoning (KIMSR) tasks. One challenge is how to effectively extract and represent rationale evidence. The current methods often extract semantically relevant but logically irrelevant evidence, resulting in flawed reasoning and inaccurate responses. We propose a two-way evidence self-alignment (TW-ESA) module, which utilizes the mutual alignment between strict reasoning and LLM reasoning to enhance its understanding of the causal logic of evidence, thereby addressing the first challenge. Another challenge is how to utilize the rationale evidence and LLM's intrinsic knowledge for accurate reasoning when the evidence contains uncertainty. We propose a dual-gated reasoning enhancement (DGR) module to gradually fuse useful knowledge of LLM within strict reasoning, which can enable the model to perform accurate reasoning by focusing on causal elements in the evidence and exhibit greater robustness. The two modules are collaboratively trained in a unified framework ESA-DGR. Extensive experiments on three diverse and challenging KIMSR datasets reveal that ESA-DGR significantly surpasses state-of-the-art LLM-based fine-tuning methods, with remarkable average improvements of 4\% in exact match (EM) and 5\% in F1 score. The implementation code is available at \href{https://anonymous.4open.science/r/ESA-DGR-2BF8}{https://anonymous.4open.science/r/ESA-DGR-2BF8}.      
\end{abstract}

\section{Introduction}
KIMSR is a task that necessitates multi-step reasoning and the retrieval of external knowledge to derive correct conclusion for complex questions. The pipeline of KIMSR can be briefly summarized as "claim-select-reason", where "claim" means to query for new evidence, "select" means to extract rationale information from the retrieved evidence, and "reason" means to make reasoning based on the evidence \cite{jhamtani-clark-2020-learning,creswell2022faithful,neves-ribeiro-etal-2022-entailment,wang2023self}. Current research fine-tunes LLMs to perform KIMSR tasks based on external evidence \cite{zhao2024seer,patil2025advancing}. However, existing methods still suffer from significant limitations.

One limitation is the misalignment between retrieved evidence and logical relevance (\textbf{semantic-logic mismatch}) (see \textbf{Appendix}~\ref{appendix:case2}). Current methods often prioritize semantic similarity, leading to logically irrelevant information that misleads reasoning and introduces hallucinations with lower-quality responses \cite{aftab2024optimizing}. Moreover, due to the incomplete or noisy nature of retrieved evidence (\textbf{uncertainty-aware hallucination})(see \textbf{Appendix}~\ref{appendix:case3}), reasoning strictly on the retrieved content \cite{lin2023over,li2025search} may not suffice to ensure robust and precise reasoning. Models may either overfit to partial evidence or fall back on memorized patterns from pretraining \cite{lin2023over}, both of which can lead to spurious reasoning outcomes. 


The limitations can be regarded as \textbf{two challenges}: (1) How to effectively extract and represent rationale evidence from retrieved evidence; (2) How to utilize the rationale evidence and LLM’s intrinsic knowledge for accurate reasoning. Since LLMs derive knowledge from associated patterns in large text corpora \cite{feng2024pre}, and have uncertainty in the reasoning of precise relationships, utilizing LLMs fine-tuning is difficult to effectively address the two challenges. Prior researches train strict models, which means that the models make decisions mainly based on the selected rationale evidence, to realize accurate reasoning \cite{zhou2024learnware}. These methods are not as effective as LLMs, because strict models lack sufficient intrinsic knowledge. Recent studies use distillation methods to transfer the knowledge of LLMs to strict models \cite{hsieh2023distilling}, but this will inject irrelevant details, thereby compromising the efficacy of evidence-based reasoning.

We design a rationale information extraction (RIE) module to train a strict model. Inspired by Mplug-owl2 \cite{ye2024mplug}, which performs multi-data feature representation alignment, we propose a \textbf{two-way evidence self-alignment (TW-ESA)} module to use the knowledge of LLMs to guide the rationale evidence extraction of strict model, and use the strict model to constraint the evidence representation of LLMs, thereby enhancing the model’s ability to deeply understand the causal relations within evidence. In addition, existing studies have verified the hidden states of LLMs contain knowledge relevant to factual judgments \cite{AzIS@2023,hu2024towards}. We design a \textbf{dual-gated reasoning enhancement (DGR)} module that can select useful knowledge representations of the model to complement external evidence, and filter out knowledge prone to hallucination. This design enhances the model's robustness when evidence exists with uncertainty and enables flexible integration of internal and external knowledge for comprehensive reasoning. The contributions are summarized as below:

\begin{itemize}
    \item To solve the problem of semantic-logic mismatch, the TW-ESA, which contains token-level and hidden-state alignments, can facilitate the guidance of rationale evidence extraction by the LLM’s intrinsic knowledge, concurrently enhancing the LLM’s capacity for evidence representation based on the constraint of strict model.
    \item To address uncertainty-aware hallucination, the DGR adopts two-layer gating mechanisms to gradually explore the optimal states for fusing the inherent knowledge of LLM within the strict model, enabling accurate reasoning despite uncertain evidence.
    \item An end-to-end learning model, named ESA-DGR, is proposed by incorporating TW-ESA and DGR into a unified framework. The model can achieve a causal mapping from rationale evidence extraction, knowledge-enhanced reasoning to the golden answer. 
\end{itemize}

\section{Related Work}
\subsection{Knowledge-Intensive Multi-Step Reasoning (KIMSR)}
KIMSR faces two challenges: (1) the lack of intermediate reasoning steps, causing logical gaps in information aggregation, (2) a lack of robustness to noise, where models may be distracted by seemingly relevant but semantically irrelevant content \cite{trivedi-etal-2023-interleaving,easily_distracted_by_irrelevant}.

To address these challenges, recent research employs Retrieval-Augmented Generation (RAG) for multi-turn retrieval and adaptive reasoning \citep{fid,Realm}. Methods like Adaptive-RAG \citep{adaptive}, Self-Ask \citep{selfask}, RA-ISF \citep{ra_isf}, and Least-to-Most \citep{least_to_most} iteratively combine retrieval and reasoning to construct answers. Search-o1 \citep{li2025search} interleaves chain-of-thought generation and document access for compositional reasoning. However, these methods lack deep discussion of the effective utility of LLM knowledge in evidential reasoning, leading to difficulties in error correction when evidence is lacking or reasoning chains are incomplete.

\subsection{Evidence Granularity and Extraction}
Evidence extraction plays a critical role in enabling LLMs to perform KIMSR. The effectiveness of RAG largely depends on the quality of the retrieved evidence. Existing approaches can be categorized into three levels: document-level (e.g., FiD~\cite{fid}, DPR~\cite{DPR}), span-level (e.g., DensePhrases~\cite{densephrase}), and compressed segment-level (e.g., SEER~\cite{zhao2024seer}). Document-level methods feed top-$k$ retrieved documents directly into the LLM, introducing irrelevant content \citep{chunkrag}. Span-level and compressed segment-level methods aim to improve input precision and compactness. For example, SEER proposes a model-based evidence extraction method trained with self-alignment using external data \citep{zhao2024seer}. RA-ISF~\citep{ra_isf} combines task decomposition with iterative relevance scoring to refine candidate evidence.

However, these methods mainly depend on static heuristics or attention mechanisms, lacking modeling between evidence selection and reasoning. Additionally, overemphasizing explicit evidence extraction might cause models to overly focus on local text cues, ignoring their inherent common sense and reducing reasoning robustness.

\subsection{Inter-model Alignment}
Alignment is an important research direction in the field of AI, aiming to ensure that the behavior of AI aligns with human expectations. As a branch of this field, inter-model alignment focuses on how to make multiple models consistent in terms of knowledge representation, reasoning approaches. MCKD designs a multi-stage collaborative distillation to make alignment between teacher model and two student models \cite{zhaoMCKD@2024}. SEER trains three expert models to realize self-alignment \cite{zhao2024seer}. PAA achieves advantage alignment between LLM agents through opponent shaping \cite{DuquePAA@2025}. mPLUG-Owl2 proposes a modal adaptive module (MAM) to achieve the alignment of visual features and language features in a shared semantic space \cite{ye2024mplug}.   

\section{Preliminary}
KIMSR aims to generate a comprehensive solution for each complex question $q$, including a reasoning chain $R$ and the final answer $a$. The model uses external knowledge sources $D$ to claim and gather necessary evidence $CE=\{C_1:E_1,...,C_n:E_n\}$. In the $i$th reasoning round, the model generates the current claim $C_i$, and searches for relevant evidence $E_i$ from $D$. Our focus is on enabling the model to effectively reason based on evidence $E_i$, leading to the correct answer $a$ or identifying missing evidence. The objective function is as follows:

\begin{equation}
\begin{aligned}
output &= \text{ESA-DGR}(E_i, CE_{<i}, q) \\
& output \in \{a,R,claim\}
\end{aligned}
\end{equation}

\noindent where ESA-DGR is the proposed model, and its input includes the retrieved evidence $E_i$ in the $i$th round reasoning, previous claims and evidence $CE{<i}$, and query $q$. The $output$ is contingent upon the reasoning progress, encompassing the reasoning chain $R$, explicit $claim$ of missing evidence, or the ultimate answer $a$.

\begin{figure*}[t]  
\centering
\includegraphics[width=\textwidth]{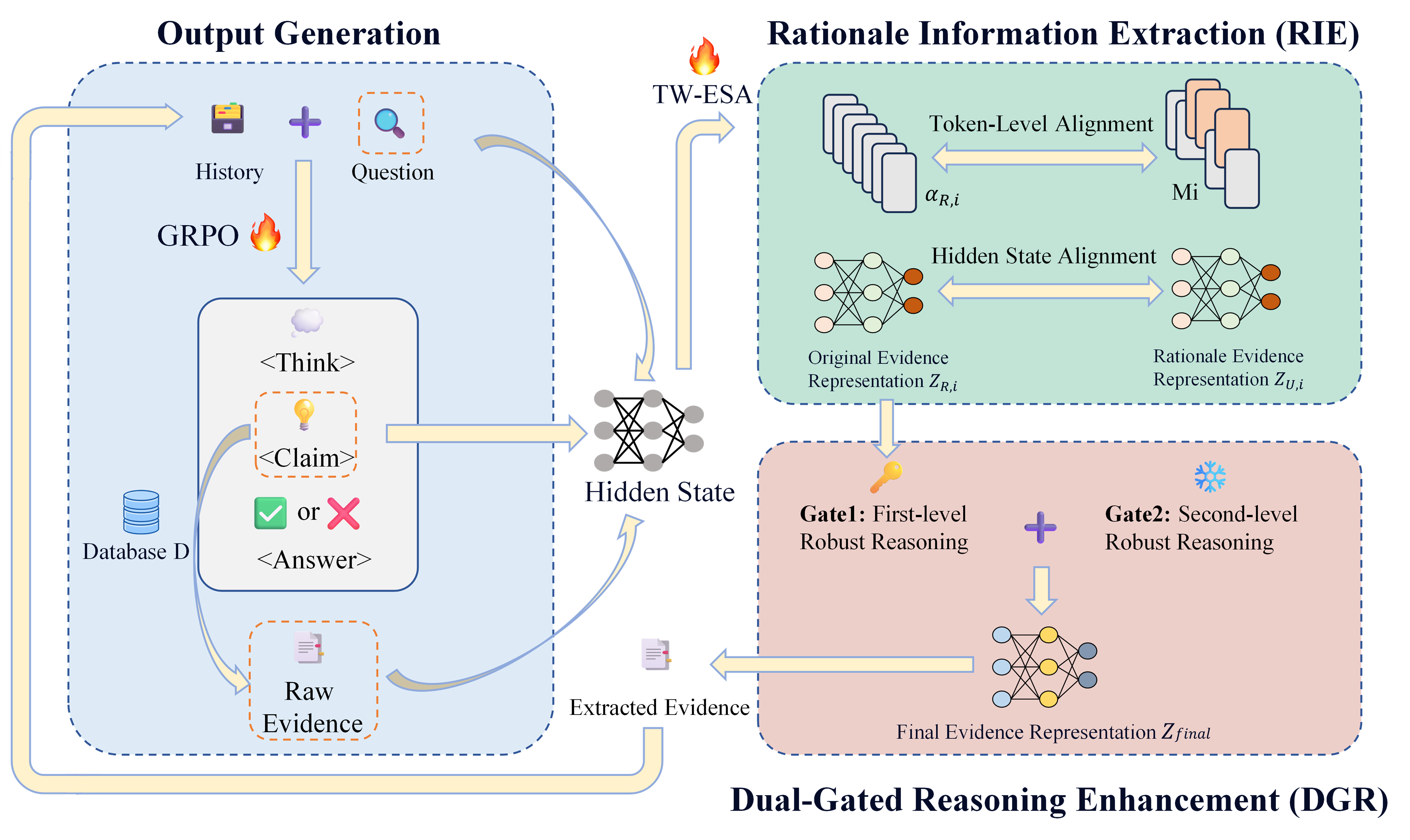}
\caption{The proposed ESA-DGR model.}
\label{fig:framwork}
\end{figure*}

\section{Methodology}
The proposed model’s framework\ref{fig:framwork} includes a \textbf{Rationale Information Extraction (RIE)} module, which can extract rationale evidence $E^{'}_{i}$ from retrieved evidence $E_{i}$, and calculate hidden states of both $E^{'}_{i}$ and $E_{i}$. The TW-ESA, encompassing both token-level and hidden-state self-alignment, is proposed to enhance RIE. We define the reasoning based on $E^{'}_{i}$ as strict reasoning. To prevent the strict model's over-reliance on $E^{'}_{i}$, a \textbf{Dual-Gating (DG)} Mechanism is proposed to integrate the knowledge of both the fine-tuned and the original LLM model, enabling it to utilize LLM's correct knowledge to refine the reasoning. Finally, \textbf{Collaborative Training (CT)} leverages GRPO to enhance the model’s self-optimization.  

\subsection{Rationale Information Extraction (RIE)}
\textbf{Token selection for reasoning. }Given a claim $C_i$ and relevant evidence $E_i$, the RIE focuses on extracting text snippets important for reasoning from $E_i$. For the $j$th token $e_j$ in $E_i$, an indicator $m_i[j]$ is assigned, where $m_i[j]=1$ means $e_j$ is selected. We adopt the hard-Kumar Distribution-based reparameterization function $k(m_i[j]|C_i, e_{<j})$ to calculate $m_i[j]$ for each token \cite{FigIRG@2019}. For all tokens in $E_i$, the indicator set is $M_i=\{m_i[1],...,m_i[|E_i|]\}$, where $m_i[j]$ is the indicator of $e_j \in E_i$, and $|E_i|$ is the number of tokens in $E_i$. The selected evidence $E^{'}_i = M_i \odot E_i$, where $\odot$ is element-wise production. 

Drawing on methods in summary generation \citep{yue2022dare}, a regularizer $\mathcal{L}_s=\lambda_1 \times \sum_{j=0}^{|E_i|}|m_i[j]| + \lambda_2 \times \sum_{j=0}^{|E_i|}|m_i[j]-m_i[j-1]|$ with respect to the selections where the first term penalizes the number of selections, and the second one encourages continuity of selections. Compared to the attention mechanism, token selection can achieve higher computational efficiency and smaller variance fluctuations, which can better fit the complex data distribution \citep{BaastDBV@2019}. 

\textbf{Hidden state representation. } Given a LLM, the hidden state of the last layer of its input can be represented as $z(q,CE_{\leq i})$, the size of which is $[1, L, d]$, where 1 is the batch size, $L$ is the length of the input (including $q$, $CE_i$) and $d$ is the dimensional length of each token. We adopt two local self-attentions $SA_{R}$ and $SA_{U}$ to calculate the hidden states on both $E_i$ and $E^{'}_{i}$:

\begin{equation}
\begin{aligned}
& Z_{\text{R},i} = f(\alpha_{\text{R},i} \odot ZV);\  \alpha_{\text{R},i}=SA_{\text{R}}(Z(q,CE_{\leq i})) \\
& Z_{\text{U},i} = f(\alpha_{\text{U},i} \odot ZV);\ \alpha_{\text{U},i}=SA_{\text{U}}(Z(q,CE^{'}_{\leq i}))
\end{aligned}
\end{equation}

\noindent where $f(.)$ is a two-layer feed-forward Network with residual connection and layer normalization. $Z_{\text{R},i}$ and $Z_{\text{U},i}$ are the hidden states of $E_i$ and $E^{'}_i$ separately; $ZV$ is the value matrix of $Z$. $\alpha_{\text{O} \in \{\text{R},\text{U}\},i}=\{\alpha_{O,i}[1], \alpha_{O,i}[2], ...\}$, where $\alpha_{O,i}[j]$ is the attention score of token $e_j$ in either $E_i$ or $E^{'}_i$.

\subsection{Two-way Self-alignment}

We define $Z_{\text{U},i}$, which uses $E^{'}_{i}$ for reasoning, as the hidden state of the strict reasoning model. To achieve mutual enhancement between $Z_{\text{R},i}$ and $Z_{\text{U},i}$, we introduce token-level and hidden-state level alignment.

\textbf{Token level alignment.}  We use cross entropy $CRE(\alpha_{\text{R},i}, M_i)$ to calculate the degree to which the distribution of $M_i$ approximates $\alpha_{\text{R},i}$. During the initial training phase, CRE can guide the training of $M_i$ with $\alpha_{\text{R},i}$, ensuring that $M_i$ focuses on semantic understanding rather than merely position and pattern matching, which can effectively prevent the extraction of meaningless tokens. 

\textbf{Hidden state alignment.} We use Jensen-Shannon divergence $JS(Z_{R,i} || Z_{U,i})$ to calculate the difference between $Z_{R,i}$ and $Z_{U,i}$ of $E_i$ and $E^{'}_i$. The reason for using $JS$ is to expect that $Z_{R,i}$ and $Z_{U,i}$ align with each other by reducing the distance between them. In this process, under the guidance of $Z_{R,i}$, $Z_{U,i}$ can fully utilize the information extracted from $E^{'}_i$. Under the constraint of $Z_{U,i}$, $Z_{R,i}$ is primed to emphasize causal reasoning information within $E_i$ pertinent to the task.

\textbf{Collaborative alignment training.} A loss function based on both $CE$ and $JS$ is proposed to realize alignment training. The purpose of alignment training is to enable RIE to enhance its ability to select and represent key information, while acquiring the existing capabilities of LLM. The loss function $\mathcal{L}_align$ could be seen as follows:

\begin{equation}
\begin{aligned}
\mathcal{L}_{align}= & \lambda_3 \times CRE(\alpha_{R,i},M_i) + \\
& \lambda_4 \times JS(Z_{R,i} || Z_{U,i}) + \lambda_5 \times \mathcal{L}_s 
\end{aligned}
\end{equation}

\noindent where $\lambda_3$, $\lambda_4$ are hyper-parameters,  $\lambda_5$ is Lagrange multiplier \cite{BaastDBV@2019}, which should be estimated. $\mathcal{L}_s$ has been introduced in Section 4.1.

\subsection{Dual-Gated Mechanism}
In order to prevent strict model $Z_{U,i}$ from overly relying on the selected evidence $E^{'}_{i}$, and there is a certain possibility that $E^{'}_{i}$ does not contain important information in original $E_i$, we integrate $Z_{R,i}$ into reasoning, because the information in $Z_{R,i}$ enhances the causal connection associated with the target question through alignment training. We define the integration of $Z_{R,i}$’s reasoning as the \textbf{first-level robust reasoning}. Finally, to mitigate the forgetting of existing knowledge that can occur during fine-tuning of $Z_{U,i}$ and $Z_{R,i}$, we incorporate the original LLM's last hidden layer representation $Z$ to facilitate \textbf{second-level robust reasoning}.  

We propose a dual-gated mechanism to realize the robust reasoning. The mechanism dynamically regulates the knowledge input to each of the three models, contingent upon the specific requirements of the target claim. For claims that can be fully explained by $E^{‘}_{i}$, it will increase the input of $Z_{U,i}$ and reduce noise interference from relevant but useless information. For information omitted in $E^{‘}_{i}$, further reasoning clues can be sought through the first-level robust reasoning. The introduction of the second-level robust reasoning can utilize the knowledge and memory of the original LLM to solve the problem of the first-level robust reasoning being overly dependent on existing evidence, such as the model repeatedly doubting why “the capital of the United States is Washington”. The formula of dual-gated mechanism can be seen as below:

\begin{equation}
\begin{aligned}
& gate1 = \sigma([Z_{U,i};Z_{R,i}] \times W_A) \\
& Z^{'}=\underbrace{gate_1 \odot Z_{U_i} + (1-gate_1) \odot Z_{R,i}}_{\text{First-level Robust Reasoning}} \\
& gate2 = \sigma([Z^{'};Z_{i}] \times W_B) \\
& Z_{final} = \underbrace{gate_2 \odot Z^{'} + (1-gate_2) \odot Z_{i}}_{\text{Second-level Robust Reasoning}}  
\end{aligned}
\end{equation} 

\noindent where ; denotes the concatenation operation. Assuming each model’s vector dimension is $d$, $W_A$ and $W_B \in \mathcal{R}^{2d \times d}$ are learnable weight matrices. The gating mechanism contains two gates $gate_1$ and $gate_2$, $gate_1$ fuses $Z_{U,i}$ and $Z_{R,i}$ to perform first-level robust reasoning and outputs the hidden layer $Z^{'}$; $gate_2$ then fuses $Z^{'}$ and $Z_i$ to perform second-level robust reasoning and outputs the final state $Z_{final}$.   

\subsection{Collaborative Training}
The robust reasoning is based on the information representation within $Z_{final}$ that maximizes the mutual information with golden answer $a$. This can be described as $max \ \text{I}(Z_\text{final};a)$. Following the auto-regression mechanism inherent in large language models (LLMs), we can express the probability $P(a|Z_{final}) = \prod_{t=1}^{T_a} P(a_t|Z_{final})$. Then, we can obtain the lower bound of $I(a; Z_{final})$ as \cite{ZhaoTG@2024}:

\begin{equation}
\begin{aligned}
I(a; Z_{final}) \geq E_{P(a,Z_{final})}[logP(a|Z_{final})] - H(a)
\end{aligned}
\end{equation}

\noindent where $H(a)$ is the entropy of answer $a$. Thus, as introduced in previous studies, maximizing the lower bound of a mutual information can be used to construct objective functions for causal association. We expect that the model can autonomously learn the retrieval, evaluation, and reasoning of evidence. Therefore, we construct an objective function $\mathcal{J}(\theta_\text{U})$ to maximize $P(a|h_u)$ based on the Group Relative Policy Optimization (GRPO) framework.   

\begin{equation}
\begin{aligned}
\mathcal{J}(\theta) \propto \frac{1}{G} \sum_{i=1}^{G} \frac{1}{T_{a,i}} \sum_{t=1}^{T_{a,i}}(\frac{\pi_{\theta}(a_{i,t}|q,CE_{<t})}{\pi_{\theta_{init}}(a_{i,t}|q,CE_{<t})})\hat{A}_i
\end{aligned}
\end{equation}

\noindent where $G$ is the number of groups, $a_i$ is the generated $i$th output, $\pi_{\theta}$ is the $Z_{final}$ model with parameter $\theta$, $\hat{A}_i$ is the output-level relative advantage of the $i$th output. The target of $\mathcal{J}(\theta)$ is to maximize $\pi_\theta$ with high $\hat{A}_i$ value, which is consistent with maximizing $P(a|Z_{final})$. We define three labels for the training process of GRPO and design corresponding format rewards (see \textbf{Appendix A.2}). The \textbf{<think>} label stores the reasoning chain \textbf{$R$}, the \textbf{<claim>} label \textbf{claims} missing evidence and simultaneously activates the retriever for evidence retrieval, and the \textbf{<answer>} label stores the final answer \textbf{$a$}. The objective $\mathcal{L}_{final}$ of the entire ESA-DGR can be seen as follows:

\begin{equation}
\begin{aligned}
\mathcal{L}_{final} = \mathcal{J}(\theta) + \mathcal{L}_{align} + D_{KL}(Z_{final} || Z^{'})
\end{aligned}
\end{equation}

\noindent where $\mathcal{L}_{align}$ is introduced in Formula (3). The purpose of using KL divergence $D_{KL}$ as the regularization term of the objective function $\mathcal{L}_{final}$ is to prevent the model parameters trained by the second-level robust reasoning from deviating too far from the first-level robust reasoning, which could lead to the loss of rigorous reasoning ability. The detailed training process of the loss function is described in\textbf{ \text{Appendix A.1}.  }

\section{Experimental Setup}

\begin{figure*}[t]
\centering
\includegraphics[width=\textwidth]{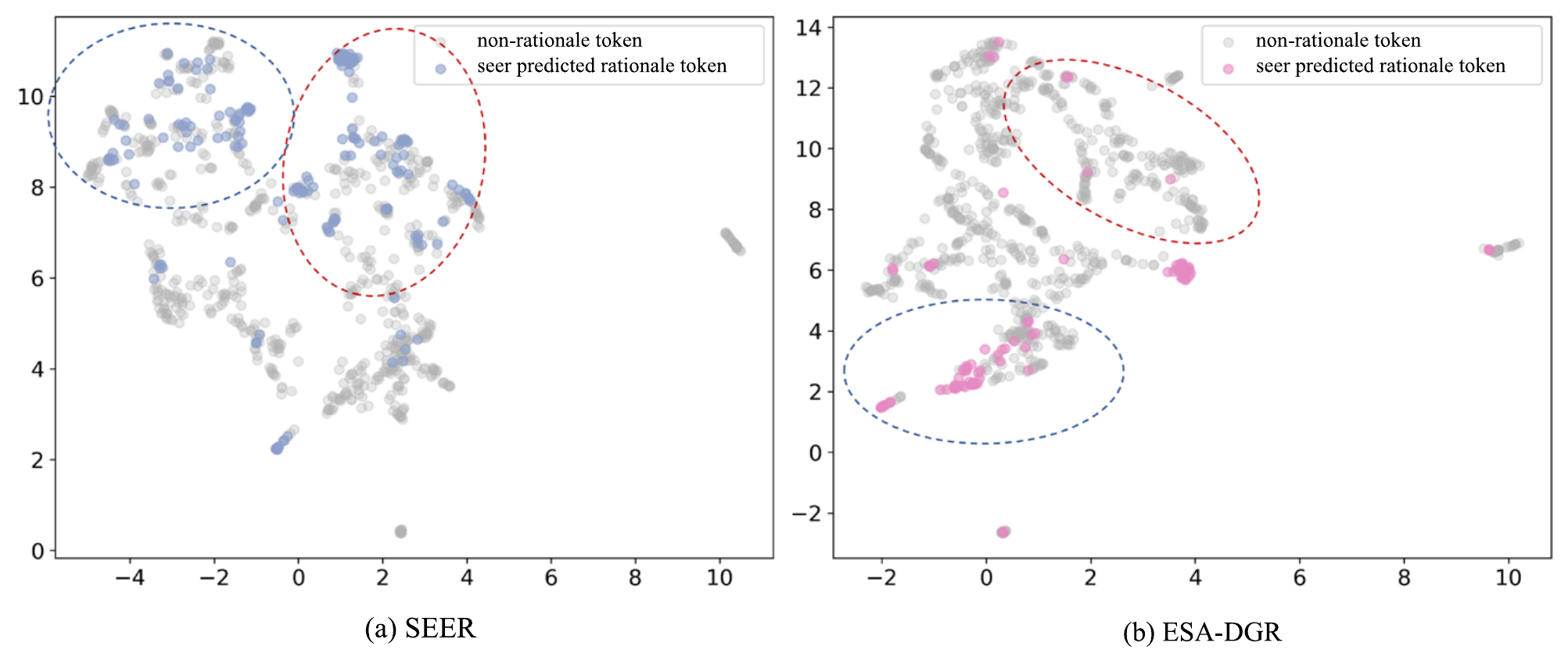}
\caption{Visualization of token representations for rationale selection. Blue: correctly predicted rationale tokens; Red: misclassified tokens by SEER (a) and TW-ESA (b). TW-ESA demonstrates better separation between rationale and non-rationale tokens. The corresponding case is detailed in \textbf{Appendix}~\ref{appendix:case1}.}
\label{fig:token}
\end{figure*}


\subsection{Datasets}

To consider complex query scenarios, we use three benchmark multi-hop QA datasets, which require sequential reasoning over multiple documents, namely \textbf{1) HotpotQA} \cite{yang2018hotpotqa},\textbf{ 2) 2WikiMultiHopQA} \cite{ho2020constructing} and\textbf{ 3) Musique }\cite{trivedi2022musique}.

\subsection{Baselines}
We compare our method with a diverse set of 10 baselines to evaluate reasoning accuracy, interpretability, and retrieval efficiency. These baselines are grouped into two major categories: (1) non-retrieval-based methods and (2) retrieval-based methods. Among retrieval-based methods, we further distinguish between those that perform rationale/evidence selection and those that do not. A detailed description of each baseline is provided in \textbf{Appendix}~\ref{appendix:baselines}.

\subsection{Implementation Details}


Following the setup in IRCoT \citet{trivedi-etal-2023-interleaving}, we construct retrieval indices for all three datasets using BM25, implemented via Elasticsearch. The retrieved top-$k$ passages are used as candidate evidence $E_i$ for each intermediate claim.

For model training, we adopt the \texttt{swift} \footnote{\url{https://github.com/modelscope/ms-swift}}framework and extend it to support joint training with alignment loss and GRPO-based policy optimization. All experiments are conducted on 8 NVIDIA A800 GPUs. Our model uses the same tokenizer and embedding initialization as the underlying LLMs (Qwen2.5-7B and LLaMA3.1-8B).

\subsection{Metrics}


We adopt both standard and auxiliary metrics to evaluate answer quality and reasoning efficiency.

For answer quality, we report \textbf{Exact Match (EM)}, \textbf{F1}, \textbf{Precision}, and \textbf{Recall}, which are standard in QA evaluation.

For auxiliary metrics, we include three indicators: (1) the \textbf{average number of retrieval queries for correctly answered examples}, denoted as \(\mathcal{Q}_{\text{avg}}\), reflecting retrieval efficiency; (2) the usefulness score of each issued claim, \textit{Uc}, rated by GPT-4.1 on a 1--5 scale. To further assess retrieval quality, we evaluate each sub-question along four dimensions: necessity~\cite{liu-etal-2022-generated}, relevance~\cite{wolfson-etal-2020-break, perez-etal-2020-unsupervised}, information gain~\cite{min-etal-2019-multi}, and reasoning progression~\cite{zhang2210automatic}. GPT-4 assigns scores using the G-EVAL protocol~\cite{liu-etal-2023-g}, and final sub-question scores are aggregated via SEER’s CoV-based weighting~\cite{zhao2024seer} and averaged across all sub-questions per original question; and (3) the \textbf{evidence score}, \(\mathcal{S}_{\text{evidence}}\), measuring the \emph{conciseness}, \emph{usefulness}, and \emph{faithfulness} of extracted evidence, following the SEER protocol~\cite{zhao2024seer}.

\section{Experiment Results}

\subsection{Main Results}
The main results of Qwen2.5-7B are presented in Table~\ref{tab:multihop_qa}, and the corresponding results on LLaMA3.1-8B can be found in\textbf{ Appendix}~\ref{sec:appendix-llama}. From the experiments, we draw the following key observations:

\textbf{ESA-DGR achieves state-of-the-art performance across all datasets.}  
Our proposed ESA-DGR framework significantly outperforms all baselines on three datasets under the Qwen2.5-7B backbone. It achieves absolute improvements of +5.5 EM / +6.1 F1 on HotPotQA and +4.3 EM / +5.7 F1 on 2WikiMultiHopQA compared to the previous best-performing method, confirming the robustness of our method in complex multi-hop reasoning.

\begin{figure}[htbp]
\centering
\includegraphics[width=0.5\textwidth]{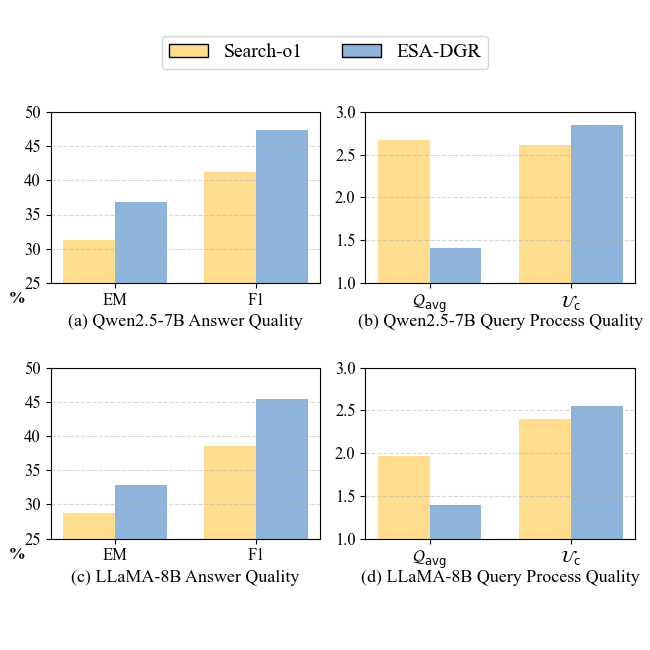}
\caption{Answer quality and query efficiency comparison between ESA-DGR and Search-o1 on Qwen2.5-7B and LLaMA-8B. ESA-DGR consistently yields better answers (EM/F1) and higher-value queries (\(\mathcal{Q}_{\text{avg}}\), \(\mathcal{U}_{\mathcal{C}}\)).}

\label{fig:ava query and usefulness}
\end{figure}

\textbf{ESA-DGR improves reasoning efficiency without additional query cost.}  
Despite its higher performance, ESA-DGR maintains a comparable or even lower number of queries compared to Search-o1. As shown in Figure~\ref{fig:ava query and usefulness}, ESA-DGR achieves high-quality answers with a lower  \(\mathcal{Q}_{\text{avg}}\), and the \textit{Uc} of each query is significantly higher. This demonstrates that ESA-DGR not only reduces unnecessary information retrieval but also prioritizes high-value queries.

\begin{figure}[htbp]
\centering
\includegraphics[width=0.5\textwidth]{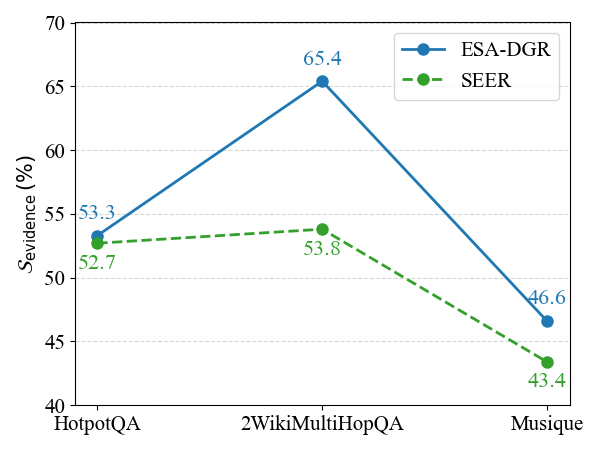}
\caption{Evidence quality comparison. ESA-DGR outperforms baseline method SEER in \(\mathcal{S}_{\text{evidence}}\), demonstrating superior rationale extraction quality.}

\label{fig:evidence}
\end{figure}

\textbf{ESA-DGR enables more accurate evidence extraction and better interpretability.}  
To assess the interpretability and evidence quality of our method, we first evaluate the extracted rationales across datasets, and find that ESA-DGR consistently achieves higher evidence quality scores (see Figure~\ref{fig:evidence}).

To further examine whether ESA-DGR has effectively disentangled the textual input into rationale and non-rationale representations, we visualize the token representations using UMAP in Figure~\ref{fig:token}. In Figure~\ref{fig:token}(a), the blue circle highlights tokens correctly predicted as rationales by SEER, while the red circle denotes tokens incorrectly selected, which lie close to the blue cluster—indicating poor separation. In contrast, Figure~\ref{fig:token}(b) shows that ESA-DGR successfully pushes the incorrectly predicted tokens (red) away from the rationale cluster (blue), suggesting better disentanglement and evidence boundary learning. This confirms ESA-DGR's advantage in separating relevant from irrelevant content during rationale selection.

\begin{table*}[ht]
\centering
\resizebox{\linewidth}{!}{%
\begin{tabular}{l|cccc|cccc|cccc}
\toprule
\textbf{Method} & \multicolumn{4}{c|}{\textbf{HotPotQA}} & \multicolumn{4}{c|}{\textbf{2WikiMultiHopQA}} & \multicolumn{4}{c}{\textbf{MuSiQue}} \\
 & EM & F1 & Prec & Recall & EM & F1 & Prec & Recall & EM & F1 & Prec & Recall \\
\midrule
\multicolumn{13}{c}{\textit{Non-Retrieval-Based Methods}} \\
\midrule
Direct & 18.2 & 26.7 & 28.3 & 27.0 & 23.4 & 28.3 & 28.4 & 28.7 & 3.4 & 9.3 & 10.9 & 9.1 \\
CoT & 19.2 & 26.3 & 28.3 & 26.0 & 23.9 & 29.8 & 29.5 & 30.7 & 4.3 & 11.6 & 12.3 & 12.1 \\
GRPO & 19.4 & 26.1 & 28.2 & 25.7 & 23.1 & 26.6 & 26.9 & 26.6 & 3.3 & 8.0 & 9.3 & 7.6 \\
\midrule
\multicolumn{13}{c}{\textit{Retrieval-Based Methods (w/o Rationale Selection)}} \\
\midrule
RAG & 23.3  & 30.6 & 32.4 & 30.6 & 21.8 & 26.3 & 26.3 & 26.8 & 3.1 & 8.9 & 10.2 & 8.7 \\
AdaptiveRAG & 21.9 & 29.1 & 30.9 & 29.4 & 22.5 & 26.6 & 26.8 & 26.6 & 3.2 & 9.0 & 10.2 & 8.6 \\
Search-o1 & \underline{31.3} & \underline{41.2} & \underline{43.0} & \underline{42.0} & \underline{45.2} & \underline{52.4} & \underline{51.8} & \underline{54.8} & \underline{8.2} & \underline{14.9} & \underline{16.2} & \underline{14.9} \\
Ra-isf & 23.4 & 31.1 & 33.0 & 31.5 & 30.3 & 32.7 & 33.1 & 32.9 & 2.5 & 7.2 & 8.5 & 7.1 \\
\midrule
\multicolumn{13}{c}{\textit{Retrieval-Based Methods (with Rationale Selection)}} \\
\midrule
DSLR & 23.9 & 31.3 & 33.2 & 31.5 & 30.2 & 34.4 & 34.3 & 35.0 & 3.1 & 6.9 & 7.9 & 6.8 \\
SEER & 27.5 & 36.6 & 38.9 & 37.0 & 34.6 & 40.1 & 39.9 & 41.0 & 3.5 & 9.0 & 9.9 & 9.1 \\
DARE & 29.0 & 38.8 & 40.5 & 39.7 & 20.9 & 31.5 & 30.2 & 35.4 & 6.1 & 12.4 & 13.1 & 13.1 \\
\midrule
ESA-DGR & \textbf{36.8} & \textbf{47.3} &\textbf{ 49.6} &\textbf{ 47.5} & \textbf{49.5} & \textbf{58.1} & \textbf{57.3} & \textbf{59.0} &\textbf{ 10.5} & \textbf{18.0} & \textbf{19.2} & \textbf{18.5} \\
\bottomrule
\end{tabular}
}
\caption{Performance comparison across three multi-hop QA datasets on Qwen2.5-7b. Baselines are categorized into non-retrieval, retrieval without rationale selection, and retrieval with rationale selection.}
\label{tab:multihop_qa}
\end{table*}


\subsection{Ablation Study}

To evaluate the role of each component in \textsc{ESA-DGR}, we conduct ablation experiments on HotpotQA using Qwen2.5-7B. Results are shown in Table~\ref{tab:ablation-hotpotqa}. The experimental results demonstrate the effectiveness of each component, especially when compared to \textbf{w/o DGR}, highlighting the importance of the dual-layer gate control in gradually fusing knowledge into the strict model.

\textbf{w/o RIE+DGR (SEER+GRPO):} Replaces our rationale extraction and gated reasoning modules with the SEER-style evidence extraction while retaining GRPO training.

\textbf{w/o TokenAlign:} Removes token-level alignment (CRE loss), which guides rationale selection via LLM attention.

\textbf{w/o StateAlign:} Disables hidden-state alignment by removing the JS divergence between $Z_{R,i}$ and $Z_{U,i}$.

\textbf{w/o DGR-Gate1:} Removes the first-level fusion between $Z_{U,i}$ and $Z_{R,i}$ in the gated reasoning module.

\textbf{w/o DGR-Gate2:} Removes the second-level fusion with the original LLM hidden state $Z$.

\textbf{w/o DGR:} Integrate $Z_{R,i}$, $Z_{U,i}$ and Z using one layer gating mechanism.

\textbf{w/o GRPO:} Replaces the GRPO training objective with standard supervised cross-entropy loss.


\begin{table}[ht]
\centering
\small
\setlength{\tabcolsep}{4pt}

\begin{tabular}{l|cccc}
\toprule
\textbf{Method Variant} & \textbf{EM} & \textbf{F1} & \textbf{Prec} & \textbf{Recall} \\
\midrule
\textbf{ESA-DGR (Full)} & \textbf{36.8} &\textbf{ 47.3} & \textbf{49.6 }& \textbf{47.5} \\
w/o RIE+DGR (SEER+GRPO) & \underline{35.0} & 45.6 & \underline{47.9} & \underline{46.0} \\
w/o TokenAlign & 32.1 & \underline{45.8} & 46.3 & 45.4 \\
w/o StateAlign & 28.2 & 35.2 & 37.2 & 35.0 \\
w/o DGR-Gate1                        & 33.8 & 44.2 & 45.2 & 42.7 \\
w/o DGR-Gate2                        & 32.9 & 42.5 & 44.0 & 41.5 \\
w/o DGR & 34.5 & 41.8 & 43.7 & 40.1 \\
w/o GRPO & 33.2 & 43.5 & 45.7 & 43.9 \\
\bottomrule
\end{tabular}
\caption{Ablation results on HotpotQA using Qwen2.5-7B.}
\label{tab:ablation-hotpotqa}
\end{table}

\subsection{Hyperparameter Sensitivity}

We analyze the impact of five key loss weights in $\mathcal{L}_{final} $: $\lambda_1$, $\lambda_2$ (token selection sparsity and continuity), $\lambda_3$, $\lambda_4$ (token/hidden-state alignment), and $\lambda_5$ (selection regularization). Each is varied individually while others are fixed.

As summarized in Figure~\ref{fig:sensitivity_qwen_hotpotqa} (see \textbf{Appendix}~\ref{sec:appendix-sensitivity}), the model shows stable performance across ranges. 

\subsection{Structural Sensitivity Analysis}

We further assess two structural settings: the number of retrieved passages ($k$) and the maximum number of claim steps. These influence evidence coverage and reasoning depth.

As shown in Table~\ref{tab:structural_sens}, the best performance occurs at $k=8$ and 3 claim steps. Fewer values limit information, while larger values introduce noise or over-claiming. The model performs robustly under different configurations with clear optimal ranges.

\section{Conclusions}
The ESA-DGR framework presents a significant improvement in the field of knowledge-intensive multi-step reasoning (KIMSR). By addressing the challenges of semantic-logic mismatch and uncertainty-aware hallucination, ESA-DGR achieves state-of-the-art performance on diverse KIMSR datasets. The innovative TW-ESA and DGR modules, along with their integration into a unified framework, facilitate accurate and robust reasoning by leveraging both external evidence and the intrinsic knowledge of LLMs.  

\section*{Limitations}
Although we have fully demonstrated the effectiveness of ESA-DGR experimentally, the underlying mechanism still require further investigation. Our subsequent plan involves theoretically exploring and innovating these mechanisms by integrating cutting-edge theoretical methods, such as confidence calculation based on stochastic processes, causal decomposition based Hidden layer feature alignment, as well as an upper bound proof for the DGR to enhance reasoning capabilities. In addition, our collaborative training strategy offers substantial potential for further enhancement. We are exploring the integration of gradient conflict theoretical methods to refine the joint training of GRPO and alignment techniques, thereby facilitating the development of more sophisticated causal reasoning models. We expect that the model can more accurately identify the reliability of the evidence, as well as assess the confidence in answering based on the evidence. Meanwhile, we also plan to conduct additional research to assess its performance in more complex scenarios, such as research Issues in Natural and Social Sciences.


\bibliography{anthology,custom}
\bibliographystyle{acl_natbib}

\appendix

\section{Model Details}
\label{sec:appendix-training}

\subsection{The Training Strategy of ESA-DGR}
We provide detailed collaborative training strategy of ESA-DGR. There are certain challenges in directly conducting collaborative training based on the loss function of formula (7). Firstly, in loss function $L_{align}$ (formula(3)), there is a dependency between the training of $CRE(\alpha_{R,i},M_i)$ and $JS(Z_{R,i} || Z_{U,i})$, because only when the quality of the tokens selected by $CRE$ is improved does the hidden layer representation of $JS$ become meaningful. Secondly, there exists a gradient conflict problem in the collaborative training of $\mathcal{J}(\theta)$ and $\mathcal{L}_{align}$, which leads to unstable model training and performance degradation. 

Therefore, we employ a phased training methodology, with the specific steps outlined as follows:

(1) In each epoch, $CRE(\alpha_{R,i},M_i)$ and $\mathcal{L}_{s}$ are trained firstly (The initial value of $\alpha_{R,i}$ is based on the calculation of original LLM).

(2) Then, we co-train $JS(Z_{R,i} || Z_{U,i})$ and $D_{KL}(Z_{final} || Z^{'})$ as regularizer for $\mathcal{J}(\theta)$ through the GRPO framework (At this point, we stop calculating the backward gradient propagation for $CRE(\alpha_{R,i},M_i)$) and $\mathcal{L}_{s}$.

We repeat the above steps until the training loss of GRPO converges. For a clearer explanation of the entire training process, we re-define the loss function as:

\begin{equation}
\begin{aligned}
\mathcal{L}_{final} = & \lambda_3 \times CE(\alpha_{R,i},M_i) + \lambda_5 \times \mathcal{L}_{s} + \\ 
& ind \times (\mathcal{J}(\theta) + \lambda_4 \times JS(Z_{R,i} || Z_{U,i}) + \\
& D_{KL}(Z_{final} || Z^{'}))
\end{aligned}
\end{equation}

\noindent where $ind$ can be seen as a switch, the initial value of which is 0. After the specified iteration of CRE training is executed, the value of ind will activated to 1, thereby activating the training of GRPO. 

\subsection{GRPO Optimization Procedure}

We provide implementation details of the GRPO optimization procedure used in our framework to supervise structured multi-step reasoning. This includes learning rate, batch size, gradient clipping, reward formulation, and iterative policy updates. Our GRPO framework consists of two components: (1) a structure-aware reward function that evaluates the format, logic, and conflict of each generated response, and (2) a policy optimization loop that iteratively updates the model using group-level advantage estimation.

The reward function is defined in Algorithm~\ref{alg:grpo-reward}, and the full policy optimization process adapted from \citet{shao2024deepseekmath} is shown in Algorithm~\ref{alg:grpo-optimize}.

\textbf{Algorithm~\ref{alg:grpo-reward}} defines a structure-aware reward function used in GRPO, which evaluates each generated output based on its format correctness, reasoning step order, and internal consistency. The reward is computed by linearly combining three rule-based checks with tunable weights $\alpha_1$, $\alpha_2$, and $\alpha_3$.







\begin{algorithm}[ht]
\caption{Structure-Aware Reward Function Used in GRPO}
\label{alg:grpo-reward}
\begin{algorithmic}[1]
\Require Completions $C = \{c_1, c_2, \dots, c_n\}$
\Ensure Rewards $R = \{r_1, r_2, \dots, r_n\}$

\For{$c \in C$}
    \State $r_1 \gets$ \Call{CheckFormat}{$c$}
    \State $r_2 \gets$ \Call{CheckOrder}{$c$}
    \State $r_3 \gets$ \Call{CheckConflict}{$c$}
    \State $r \gets \alpha_1 r_1 + \alpha_2 r_2 + \alpha_3 r_3$
    \State Append $r$ to $R$
\EndFor
\State \Return $R$

\Function{CheckFormat}{$c$}
    \State \Return $1$ if all tags (\texttt{<think>}, \texttt{<claim>}, \texttt{<answer>}) are present and properly closed; else $-1$
\EndFunction

\Function{CheckOrder}{$c$}
    \State \Return $1$ if tag order satisfies \texttt{<think>} $\prec$ \texttt{<claim>} $\prec$ \texttt{<answer>}; else $-1$
\EndFunction

\Function{CheckConflict}{$c$}
    \State Extract contents $s_{\text{claim}}, s_{\text{answer}}$
    \State \Return $-1$ if both non-empty; else $1$
\EndFunction

\end{algorithmic}
\end{algorithm}
\textbf{Algorithm~\ref{alg:grpo-optimize}} describes the iterative GRPO process. It alternates between generating multiple completions per prompt, scoring them using the structure-aware reward function, estimating group-level advantages, and updating the policy model to maximize expected relative advantage.

\begin{algorithm}[ht]
\caption{Iterative Group Relative Policy Optimization}
\label{alg:grpo-optimize}
\begin{algorithmic}[1]
\Require Initial model $\pi_{\theta_0}$, reward model $r_\phi$, task prompts $\mathcal{D}$; hyperparameters $\epsilon$, $\beta$, $\mu$

\State $\pi_\theta \gets \pi_{\theta_0}$
\State $\pi_{\text{ref}} \gets \pi_\theta$

\For{iteration = $1$ to $I$}
    \For{step = $1$ to $M$}
        \State Sample mini-batch $\mathcal{D}_b \subset \mathcal{D}$
        \State $\pi_{\text{old}} \gets \pi_\theta$
        \State Generate $G$ completions $\{o_i\}_{i=1}^G \sim \pi_{\text{old}}( \cdot \mid q)$ for each $q \in \mathcal{D}_b$
        \State Evaluate rewards $\{r_i\}_{i=1}^G$ using Algorithm~\ref{alg:grpo-reward}
        \State Estimate token-level advantages $\hat{A}_{i,t}$ via group-level comparison
        \For{GRPO iteration $j = 1$ to $\mu$}
            \State Update $\pi_\theta$ to maximize:
            \[
                \mathcal{J}(\theta) = \frac{1}{G} \sum_{i=1}^{G} \frac{1}{T_i} \sum_{t=1}^{T_i}
                \left( \frac{\pi_\theta(a_{i,t})}{\pi_{\text{old}}(a_{i,t})} \right) \cdot \hat{A}_{i,t}
            \]
        \EndFor
        \State Optionally update $r_\phi$ via replay buffer
    \EndFor
\EndFor
\State \Return $\pi_\theta$
\end{algorithmic}
\end{algorithm}

\begin{table*}[ht]
\centering
\resizebox{\linewidth}{!}{%
\begin{tabular}{l|cccc|cccc|cccc}
\toprule
\textbf{Method} & \multicolumn{4}{c|}{\textbf{HotPotQA}} & \multicolumn{4}{c|}{\textbf{2WikiMultiHopQA}} & \multicolumn{4}{c}{\textbf{MuSiQue}} \\
 & EM & F1 & Prec & Recall & EM & F1 & Prec & Recall & EM & F1 & Prec & Recall \\
\midrule
\multicolumn{13}{c}{\textit{Non-Retrieval-Based Methods}} \\
\midrule
Direct & 7.7 & 10.0 & 10.2 & 10.9 & 6.8 & 8.0 & 7.8 & 8.4 & 1.1 & 2.9 & 1.7 & 18.4  \\
CoT & 20.9 & 29.2 & 30.3 & 30.0 & 18.6 & 23.5 & 23.1 & 25.2 & 4.9 & 11.0 & 11.6 & 11.4 \\
GRPO & 19.4 & 26.2 & 28.2 & 25.8 & 16.3 & 22.9 & 22.8 & 23.1 & 3.1 & 9.0 & 10.8 & 8.5 \\
\midrule
\multicolumn{13}{c}{\textit{Retrieval-Based Methods (w/o Rationale Selection)}} \\
\midrule
RAG & 20.6 & 30.5 & 31.0 & 336.3 & 14.7 & 25.5 & 23.6 & 35.5 & 2.2 & 6.8 & 7.1 & 9.2  \\
AdaptiveRAG & 17.1 & 26.5 & 25.5 & 28.0 & 12.1 & 22.1 & 18.7 & 28.4 & 1.9 & 7.8 & 5.4 & 15.2  \\
Search-o1 & \underline{28.7} & \underline{38.6} & \underline{40.4} & \underline{39.6} & \underline{37.1} & \underline{43.9} & \underline{43.1} & \underline{47.2} & \underline{11.2} & \underline{18.4} & \underline{19.4} & \underline{19.0}  \\
Ra-isf& 26.6 & 35.9 & 37.8 & 39.7 & 25.5 & 31.5 & 31.3 & 38.2 & 1.9 & 7.3 & 8.9 & 13.9 \\
\midrule
\multicolumn{13}{c}{\textit{Retrieval-Based Methods (with Rationale Selection)}} \\
\midrule
DSLR & 23.9 & 31.3 & 33.2 & 31.5 & 30.2 & 34.4 & 34.3 & 35.0 & 3.1 & 6.9 & 7.9 & 6.8 \\
SEER & 17.5 & 24.0 & 24.8 & 25.8 & 21.0 & 26.0 & 25.4 & 27.5 & 2.1 & 4.9 & 5.1 & 6.0  \\
DARE & 24.0 & 31.3 & 31.6 & 34.6 & 15.3 & 20.7 & 20.5 & 21.9 & 2.6 & 7.6 & 7.8 & 8.6  \\
\midrule
ESA-DGR & \textbf{32.8} & \textbf{45.4} & \textbf{44.8} & \textbf{46.2} & \textbf{43.6} & \textbf{50.9} & \textbf{47.0} & \textbf{55.7} &\textbf{ 12.7 }& \textbf{23.6} & \textbf{25.1} &\textbf{ 22.3}  \\
\bottomrule
\end{tabular}
}
\caption{Performance comparison across three multi-hop QA datasets on Llama3.1-8b. Baselines are categorized into non-retrieval, retrieval without rationale selection, and retrieval with rationale selection.}
\label{tab:multihop_qa_llama}
\end{table*}

\begin{figure*}[t]  
    \centering
    \includegraphics[width=0.9\textwidth, keepaspectratio]{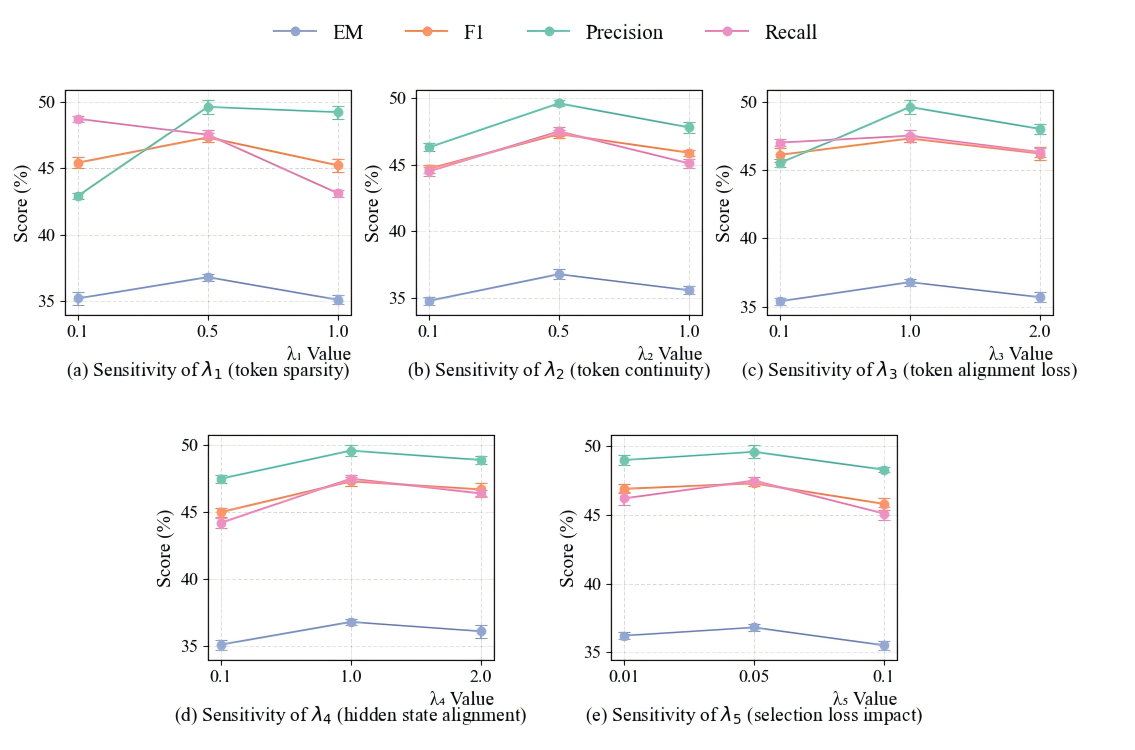}
    \caption{Sensitivity analysis of five loss-related hyperparameters ($\lambda_1$ to $\lambda_5$) on Qwen2.5-7B using the HotpotQA dataset. Each subplot shows the impact of one hyperparameter on answer accuracy (EM/F1) and reasoning faithfulness (Precision/Recall), demonstrating that ESA-DGR achieves stable performance under a range of settings and peaks consistently around $\lambda=0.5$ or $1.0$.}

    \label{fig:sensitivity_qwen_hotpotqa}
\end{figure*}

\begin{table}[ht]
\centering
\caption{Sensitivity analysis of structural parameters on HotpotQA.}
\label{tab:structural_sens}

\vspace{0.8em}
\textbf{(a) Sensitivity to number of retrieved passages (Top-k)}\\
\begin{tabular}{c|cccc}
\toprule
Top-k Docs & EM & F1 & Precision & Recall \\
\midrule
2 & 33.2 & 43.0 & 45.2 & 41.3 \\
4 & 35.1 & 45.2 & 47.1 & 44.0 \\
6 & 36.2 & 46.6 & 48.8 & 46.5 \\
\textbf{8} & \textbf{36.8} & \textbf{47.3} & \textbf{49.6} & \textbf{47.5} \\
10 & 36.5 & 46.9 & 49.0 & 47.0 \\
\bottomrule
\end{tabular}

\vspace{1.2em}
\textbf{(b) Sensitivity to maximum number of claim steps allowed}\\
\begin{tabular}{c|cccc}
\toprule
Max Steps & EM & F1 & Precision & Recall \\
\midrule
2 & 36.4 & 46.9 & 49.2 & 46.6 \\
\textbf{3} & \textbf{36.8} & \textbf{47.3} & \textbf{49.6} & \textbf{47.5} \\
5 & 36.6 & 47.0 & 49.1 & 47.1 \\
7 & 36.2 & 46.4 & 48.4 & 46.2 \\
10 & 35.7 & 45.7 & 47.2 & 45.0 \\
\bottomrule
\end{tabular}
\end{table}

\section{Baseline Details}
\label{appendix:baselines}
We describe the settings for all baseline methods. 

\textbf{Direct Prompting and Vanilla LM} \cite{brown2020language} represents a simple setting where the question is directly presented to the LLM to generate an answer, without access to any retrieved documents or reasoning steps.

\textbf{Chain-of-Thought (CoT)} \cite{wei2022chain} introduces step-by-step reasoning through prompting, encouraging the model to think before answering. 

\textbf{GRPO} \cite{shao2024deepseekmath} is a reinforcement learning baseline that optimizes the model response based on a structured reward function. 

\textbf{RAG} \cite{guu2020retrieval} retrieves a fixed number of top-ranked documents and concatenates them as input to the LLM, which then attempts to answer the question. 

\textbf{DSLR} \cite{hwang-etal-2024-dslr} improves RAG by sentence-level filtering, removing irrelevant sentences and reconstructing coherent passages (evidence) before feeding them to the model, to enhance answer relevance and conciseness.

\textbf{Search-o1} \cite{li2025search} equips the LLM with an agentic retrieval mechanism and a reasoning-in-documents module. 

\textbf{SEER} \cite{zhao2024seer} proposes a self-aligned learning framework to train an evidence extractor that selects informative and concise spans from retrieved passages. 

\textbf{RA-ISF} \cite{ra_isf} combines task decomposition and retrieval relevance feedback in an iterative loop. It allows the model to determine whether to retrieve, what to retrieve, and when to decompose the query into smaller sub-tasks.

\textbf{Adaptive-RAG} \cite{adaptive} introduces a query complexity-aware strategy. A trained classifier first predicts the complexity level of the question, then dynamically selects a reasoning strategy—ranging from no-retrieval to multi-hop retrieval—based on the prediction.

\textbf{DARE}~\cite{yue2022dare} enhances rationale extraction by disentangling input into rationale and non-rationale parts, and minimizing their mutual information to improve interpretability. It introduces CLUB-NCE, a novel mutual information estimator, and outperforms classical selector-predictor models by leveraging information from both selected and non-selected tokens.

\section{Results with LLaMA3.1-8B}
\label{sec:appendix-llama}
To verify generality across backbone models, we replicate all experiments on LLaMA3.1-8B. As shown in Table~\ref{tab:multihop_qa_llama}, ESA-DGR consistently outperforms other methods across EM and F1.

\section{Sensitivity Analysis Results}
\label{sec:appendix-sensitivity}
We report detailed results of all sensitivity analysis experiments on HotpotQA.

\vspace{0.5em}
\noindent
\textit{Observation from Table 4:} Performance improves with more retrieved passages, peaking at $k=8$ before slightly dropping, indicating a saturation point. Likewise, limiting the maximum claim steps to 3 yields the best performance, while further increasing the step count brings marginal gains or even slight drops, suggesting potential over-reasoning or noise accumulation.


\section{Case Study: Evidence Extraction Comparison}
\label{appendix:case1}
\vspace{0.5em}
\begin{tcolorbox}[colback=gray!5, colframe=black!30, title=Case Overview, fonttitle=\bfseries, boxrule=0.5pt, sharp corners=south, breakable]

\textbf{Main Question:} \\
{\small
Since when has the automobile driven by Garrett Smithley for MBM Motorsports been sold?
}

\vspace{1em}

\textbf{Claim (Sub-question):} \\[-0.5em]  
\begin{tcolorbox}[colback=yellow!10, colframe=yellow!50!black, boxrule=0.3pt, sharp corners, left=1mm, right=1mm, top=0.5mm, bottom=0.5mm, breakable]
\small
What car does Garrett Smithley drive for MBM Motorsports?
\end{tcolorbox}

\vspace{1em}

\textbf{Raw Evidence (Excerpted):} \\
{\small
\begin{spacing}{1.1}
Garrett Smithley (born April 27, 1992) is an American professional stock car racing driver. He currently competes full-time in the NASCAR Xfinity Series, driving the No. 0 Chevrolet Camaro for JD Motorsports and the No. 40 Toyota Camry for MBM Motorsports. Smithley has also competed in the Camping World Truck Series and ARCA Racing Series. [...]

The MBM Tourismo was a very low-production (probably only two were built) automobile sold by Peter Monteverdi. Monteverdi's small company MBM (standing for Monteverdi Binningen Motors) mainly focused on competition, but a "few" sports cars were also produced. [...]
\end{spacing}
}
\end{tcolorbox}

\begin{tcolorbox}[colback=white, colframe=black!40, title=Evidence Selected by Different Systems, fonttitle=\bfseries, boxrule=0.5pt, sharp corners=south, breakable]

\begin{tcolorbox}[colback=blue!2!white, colframe=blue!40!black, title=SEER Extracted Evidence, boxrule=0.5pt, sharp corners=south, breakable]
\small
The MBM Tourismo was a very low-production (probably only two were built) automobile sold by Peter Monteverdi.
\end{tcolorbox}

\vspace{0.8em}

\begin{tcolorbox}[colback=green!2!white, colframe=green!40!black, title=RIE Extracted Evidence, boxrule=0.5pt, sharp corners=south, breakable]
\small
Garrett Smithley drives the No. 0 Chevrolet Camaro for JD Motorsports and the No. 40 Toyota Camry for MBM Motorsports.
\end{tcolorbox}

\end{tcolorbox}

\section{Case Study: Semantic-Logic Mismatch}
\label{appendix:case2}

\vspace{0.5em}
\begin{tcolorbox}[colback=gray!5, colframe=black!30, title=Main Question, fonttitle=\bfseries, boxrule=0.5pt, sharp corners=south, breakable]
\textbf{Main Question:} \\
{\small
What year was the actor that co-starred with Sidney Poitier in Lieel Nikita born?
}
\end{tcolorbox}

\vspace{1em}

\begin{tcolorbox}[colback=blue!5!white, colframe=blue!30!black, title=Stage 1: Initial Sub-query and Interpretation, fonttitle=\bfseries, boxrule=0.5pt, sharp corners=south, breakable]

\textbf{Sub-query 1:} \\[-0.5em]
\begin{tcolorbox}[colback=yellow!10, colframe=yellow!50!black, boxrule=0.3pt, sharp corners, left=1mm, right=1mm, top=0.5mm, bottom=0.5mm, breakable]
\small
Sidney Poitier movies Lieel Nikita
\end{tcolorbox}

\vspace{0.8em}
\textbf{Retrieved Evidence:} \\
{\small
\begin{spacing}{1.1}
The query "Sidney Poitier Lieel Nikita movie" does not directly match any information in the provided content. However, the relevant details are:

1.Sidney Poitier is a Bahamian-American actor and filmmaker (born 1927), but no connection to a movie titled "Lieel Nikita" is mentioned.

2.The content includes unrelated references to "Nikita," such as a Ukrainian pop duo and a village in Cyprus, but none are linked to Sidney Poitier or a film.

3.There is no mention of a movie involving Sidney Poitier and the term "Lieel Nikita" in the given text.

No further relevant information exists in the source material.
\end{spacing}
}

\vspace{0.8em}
\textbf{Model's Reasoning:} \\
{\small
The search result indicates that "Lieel Nikita" might be a misspelling of "Lilies of the Field," where Sidney Poitier co-starred with Lilia Skala. However, the question asks for the actor (male) who co-starred with him. Since Lilia Skala is an actress, perhaps there's another co-star. Let me verify the cast of "Lilies of the Field."
}

\vspace{0.8em}
\textbf{Model's Acquired Knowledge:} \\
{\small
No valid result was returned.
The model did not handle the misspelling and directly inferred it as a mistaken reference to Lilies of the Field.
}
\end{tcolorbox}

\vspace{1em}

\begin{tcolorbox}[colback=blue!5!white, colframe=blue!30!black, title=Stage 2: Further Investigation, fonttitle=\bfseries, boxrule=0.5pt, sharp corners=south, breakable]

\textbf{Sub-query 2:} \\[-0.5em]
\begin{tcolorbox}[colback=yellow!10, colframe=yellow!50!black, boxrule=0.3pt, sharp corners, left=1mm, right=1mm, top=0.5mm, bottom=0.5mm, breakable]
\small
Lilies of the Field 1963 cast
\end{tcolorbox}

\vspace{0.8em}
\textbf{Retrieved Evidence:} \\
{\small
\begin{spacing}{1.1}
The 1963 film Lilies of the Field stars Sidney Poitier, Lilia Skala, Stanley Adams, and Dan Frazer. It was produced and directed by Ralph Nelson and adapted by James Poe from William Edmund Barrett's 1962 novel. The film features an early score by composer Jerry Goldsmith. Notably, it is unrelated to other films or works sharing the same title. A 1979 TV sequel, Christmas Lilies of the Field, features a different cast, including Billy Dee Williams and Maria Schell.
\end{spacing}
}

\vspace{0.8em}
\textbf{Model's Reasoning:} \\
{\small
Lilia Skala is female → model reasons “perhaps user meant a male co-star”
}

\vspace{0.8em}
\textbf{Model's Acquired Knowledge:} \\
{\small
Retrieved cast members: Lilia Skala, Stanley Adams, Dan Frazer.
}
\end{tcolorbox}

\vspace{1em}

\begin{tcolorbox}[colback=gray!5, colframe=black!30, title=Final Answer, fonttitle=\bfseries, boxrule=0.5pt, sharp corners=south, breakable]

\textbf{Conclusion:} \\
\begin{tcolorbox}[colback=red!5!white, colframe=red!50!black, boxrule=0.3pt, sharp corners, left=1mm, right=1mm, top=0.5mm, bottom=0.5mm, breakable]
\small
Incorrect answer: Stanley Adams (b. 1915)
\end{tcolorbox}
\end{tcolorbox}

\section{Case Study: Uncertainty-Aware Hallucination}
\label{appendix:case3}

\vspace{0.5em}
\begin{tcolorbox}[colback=gray!5, colframe=black!30, title=Query and Interpretation, fonttitle=\bfseries, boxrule=0.5pt, sharp corners=south, breakable]
\textbf{Question:} \\[-0.5em]
\begin{tcolorbox}[colback=yellow!10, colframe=yellow!50!black, boxrule=0.3pt, sharp corners, left=1mm, right=1mm, top=0.5mm, bottom=0.5mm]
\small
In which city is the ambassador of the Rabat-Salé-Kénitra administrative region to China based?
\end{tcolorbox}

\vspace{0.8em}
\textbf{Retrieved Evidence:} \\
{\small
\begin{spacing}{1.1}
Rabat-Salé-Kénitra is one of the twelve administrative regions of Morocco. It is situated in north-western Morocco and has a population of 4,580,866 (2014 census). The capital is Rabat.
\end{spacing}
}

\vspace{0.8em}
\textbf{Model's Reasoning:} \\
{\small
The ambassador of the Rabat-Salé-Kénitra administrative region to China would be a national ambassador, and ambassadors from Morocco to other countries are typically based in the capital city, which is Rabat. The provided reference confirms that Rabat is the capital of both Morocco and the Rabat-Salé-Kénitra region.
}

\end{tcolorbox}

\begin{tcolorbox}[colback=gray!5, colframe=black!30, title=Final Answer, fonttitle=\bfseries, boxrule=0.5pt, sharp corners=south, breakable]

\textbf{Conclusion:} \\
\begin{tcolorbox}[colback=red!5!white, colframe=red!50!black, boxrule=0.3pt, sharp corners, left=1mm, right=1mm, top=0.5mm, bottom=0.5mm, breakable]
\small
Incorrect answer: Rabat.
\end{tcolorbox}
\end{tcolorbox}

\end{document}